\documentclass[conference, letterpaper]{IEEEtran}
\usepackage[utf8]{inputenc}
\usepackage[T1]{fontenc}
\usepackage{fixltx2e}
\usepackage{graphicx}
\usepackage{grffile}
\usepackage{longtable}
\usepackage{wrapfig}
\usepackage{rotating}
\usepackage[normalem]{ulem}
\usepackage{amsmath}
\usepackage{textcomp}
\usepackage{amssymb}
\usepackage{capt-of}
\usepackage{hyperref}
\usepackage{url}
\usepackage{float}
\usepackage{gensymb}
\usepackage{hyperref}
\hypersetup{colorlinks,breaklinks,linkcolor=blue,urlcolor=blue,anchorcolor=blue,citecolor=blue}
\usepackage{tikz}
\usepackage{tabularx}
\usepackage{array}
\usepackage{FTC}
\usepackage{Authors}
\date{}
\title{Personal Food Computer: A new device for controlled-environment agriculture}
\hypersetup{
 pdfauthor={eddie},
 pdftitle={Personal Food Computer: A new device for controlled-environment agriculture},
 pdfkeywords={},
 pdfsubject={},
 pdfcreator={Emacs 25.1.1 (Org mode 8.3.6)}, 
 pdflang={English}}
\begin{document}

\maketitle
\newcommand*\circled[1]{\tikz[baseline=(char.base)]{
            \node[shape=circle,draw,inner sep=1pt] (char) {#1};}}

\newcolumntype{C}[1]{>{\centering\let\newline\\\arraybackslash\hspace{0pt}}m{#1}} 

\begin{abstract}

Due to their interdisciplinary nature, devices for
controlled-environment agriculture have the possibility to turn into
ideal tools not only to conduct research on plant phenology but also
to create curricula in a wide range of disciplines.
Controlled-environment devices are increasing their functionalities as
well as improving their accessibility. Traditionally, building one of
these devices from scratch implies knowledge in fields such as
mechanical engineering, digital electronics, programming, and energy
management. However, the requirements of an effective
controlled-environment device for personal use brings new constraints
and challenges. This paper presents the OpenAg\texttrademark{}
Personal Food Computer (PFC); a low cost desktop size platform, which
not only targets plant phenology researchers but also hobbyists,
makers, and teachers from elementary to high-school levels (K-12). The
PFC is completely open-source and it is intended to become a tool that
can be used for collective data sharing and plant growth analysis.
Thanks to its modular design, the PFC can be used in a large spectrum
of activities.

\end{abstract}


\IEEEpeerreviewmaketitle

\section{Introduction}
\label{sec:orgheadline1}
\label{orgtarget1}

In the coming decades, it is expected that humanity will need to
double the quantity of food, fiber, and fuel produced to meet global
demands. However, growing seasons are predicted to become more
volatile, and arable land (80\% is already being used) is expected to
significantly decrease, due to global warming \cite{FAO2016}.
Concurrently, public and private institutions are starting to take
an increasing interest in producing specific compounds and chemical
elements using innovative agricultural platforms (e.g.,
controlled-environment devices) for several applications (medical,
cosmetics, environmental, etc.) \cite{Lee2016,Fox2006,Olinger2012}.

In the last years, the field of phenomics \cite{Furbank2011} has
provided key insights about how different organisms can be optimized
under certain environmental conditions. These observations have
uncovered the possibility of creating ``climate recipes''
\cite{Harper2015} for specific organisms where a certain trait (e.g.,
volume, taste, chemical concentration, etc.) is maximized. The
creation and optimization of such ``recipes'' can increase the
production of valuable compounds as well as improve crop yield,
which still represents a state-of-the-art problem in the field of
modern agriculture.

\begin{figure}[!tbh]
\centering
\includegraphics[width=0.99\columnwidth]{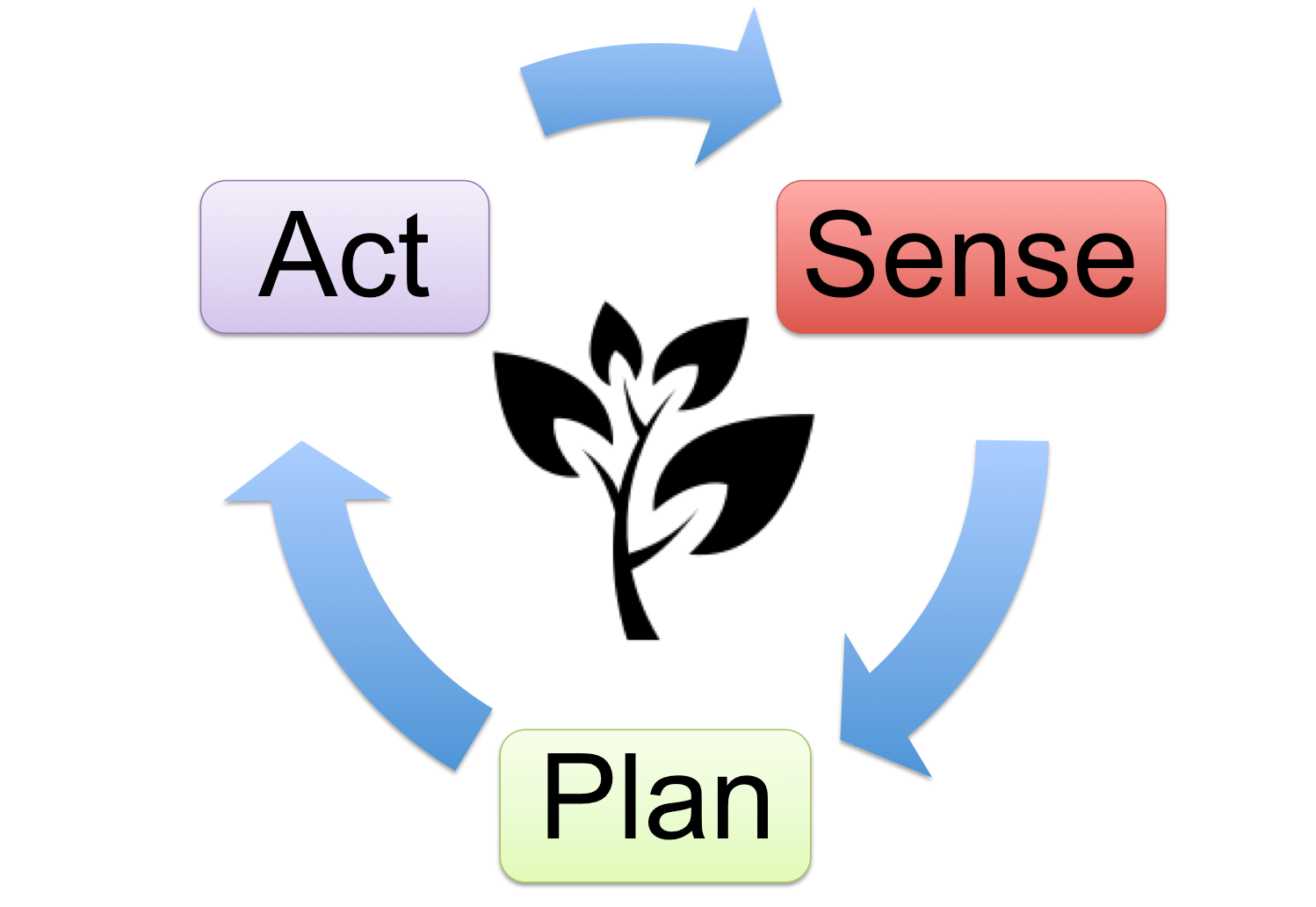}
\caption{\label{fig:orgparagraph1}
Classical feedback control loop extensively used in robotics research. This loop is composed of three main phases (Sense, Plan, Act), which in combination with modern controlled-environment agricultural devices allow treating plants as ``robotic'' agents.}
\end{figure}

Furthermore, a new framework whereby plants and crops are
controlled and monitored by computer-based algorithms has emerged
recently in the precision, and cellular agriculture fields
\cite{Abdullah2016,Zhou2016}. Agriculture Cyber-Physical Systems
(ACPS) provide the possibility not only to replicate experiments
easily, but also to collect, analyze, and learn from the data
obtained to discover new traits and patterns. ACPS have the
potential to assist plant optimization methods achieving more
autonomous, efficient, and intelligent plant growth models through
the integration of robotic control loops (Fig.
\ref{fig:orgparagraph1}) into novel agricultural devices for both
indoor and outdoor environments.

Currently, controlled-environment platforms are either based on
open-source and open-hardware standards or they have the capability
to precisely control the environment around plants and other
organisms. Moreover, very few of these devices have a suitable size for
operating outside a fully-equipped research lab. Therefore,
hindering the adoption of this technology by other user profiles.
However, the controlled-environment devices that fulfill these size
constraints suffer from a lack of customizability capabilities due
to proprietary hardware and software solutions.

The OpenAg\texttrademark{} Personal Food Computer is the first to
fulfill these characteristics in the same device. In addition, the
OpenAg\texttrademark{} Personal Food Computer allows its user to
create, store, and share the data generated during the growth cycle.
Therefore, providing the possibility of creating ``climate recipes''
and allowing other suitable devices to recreate the same
environmental conditions, improving the reproducibility of the
experiments.

\begin{figure}[!tb]
\centering
\includegraphics[width=\columnwidth]{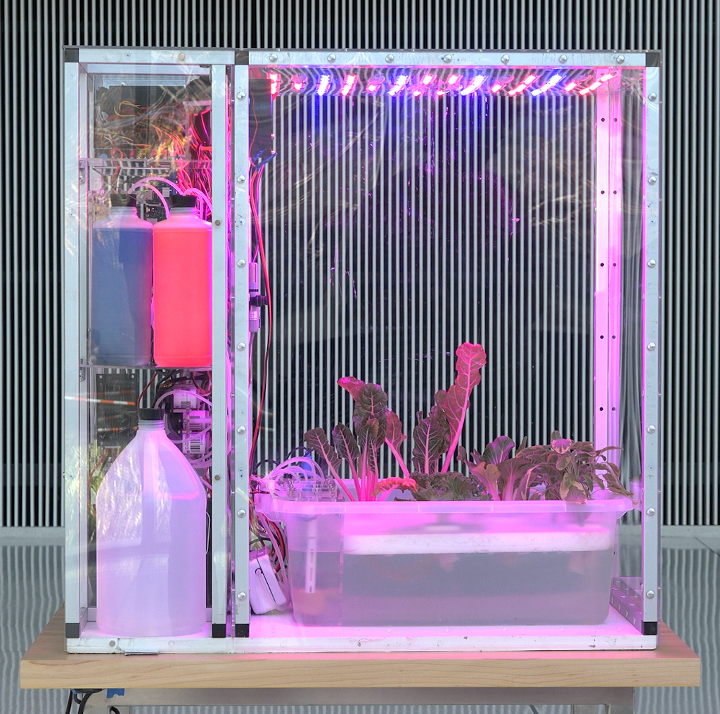}
\caption{\label{fig:orgparagraph2}
OpenAg\texttrademark{} Personal Food Computer (PFC) v2. Compared to previous versions (v1), v2 has an improved software controller and additional sensor and actuation capabilities.}
\end{figure}

This paper presents the design approach resulting in the
OpenAg\texttrademark{} Personal Food Computer named throughout the
paper as PFC (depicted in Fig. \ref{fig:orgparagraph2}), a desktop size
controlled-environment device developed at the Open Agriculture
Initiative\footnote{\url{http://openag.media.mit.edu/}} at the Massachusetts Institute of Technology
(MIT) for a wide range of activities. The main objectives of this
development were:

\begin{itemize}
\item To explore the synergy between state-of-the-art robotics methods
and controlled-environment devices to discover, analyze, and
integrate new techniques to improve plant growth models.

\item To use a personal controlled-environment device to generate,
share, and reproduce ``climate recipes'' among a community of users.

\item To use a personal controlled-environment device to produce
academic curricula from elementary to high school levels (K-12).

\(\newline\)
\end{itemize}

\section{Existing robots for personal agriculture}
\label{sec:orgheadline7}

\begin{table*}[htb]
\centering
\begin{tabular}{|C{2.15cm}|C{2.15cm}|C{2.15cm}|C{2.15cm}|C{2.15cm}|C{2.15cm}|C{2.15cm}|}
\hline
Characteristics & Farmbot & AeroGarden & Leaf & Grove & Conviron A1000 & PFC\\
\hline
Target Consumer & Consumer & Consumer & Consumer & Consumer & Scientist & Consumer\\
Price & \$3100 & \$99-\$379 & \$1500 & \$4500 & On request & \$3000\footnotemark\\
Platform & Open & Closed & Closed & Closed & Closed & Open\\
Adaptive Env. & No & No & Controlled & No & Controlled & Controlled\\
Customizability & High & Low & Low & Low & Low & High\\
\hline
\end{tabular}
\caption{\label{tab:orgtable1}
Comparison of several agricultural devices for personal and scientific use.}

\end{table*}\footnotetext[2]{Total cost of all needed components described in section \ref{orgtarget2}.}

The number of agricultural devices available on the market is
increasing. In this section, we summarize a small subset of them
that can be used as personal platforms. Table
\ref{tab:orgtable1} summarizes the main characteristics of the
devices and technologies cited in this section.

\subsection{Farmbot}
\label{sec:orgheadline2}
\label{orgtarget3}

The Farmbot\footnote{\url{http://farmbot.io/}} is an open-source Computer Numeric Control
(CNC) machine that allows the user to plant small herbs and
vegetables in an outdoor 2D grid layout (4.5 \(m^{2}\) or 14.7
\(ft^{2}\)). Optimized for backyard usage, the Farmbot can perform
operations such as watering, spraying, and seed spacing with a
single end effector due to its exchangeable head tool. Even though
it incorporates data acquisition and analysis tools, the Farmbot
cannot control the environment since it is designed for outdoor
use. On the other hand, the Farmbot is open source, fully
documented and customizable. 

\subsection{AeroGarden}
\label{sec:orgheadline3}
\label{orgtarget4}

The AeroGarden platform\footnote{\url{http://www.aerogarden.com/}} is a consumer kit for
growing herbs, small flowers, and plants. Ranging from \$99-\$379 in
price, AeroGarden devices provide enhanced capabilities such as
WiFi connectivity to smartphones or 45 Watts of LED lighting.
However, AeroGarden is a closed platform. In addition, the
environment is uncontrolled and affected by the surrounding
climate. Finally, the customizability of the system is low and its
inputs are proprietary. 

\subsection{Leaf}
\label{sec:orgheadline4}
\label{orgtarget5}

The Leaf platform\footnote{\url{http://www.getleaf.co/}} is a medium-size (600 \(\times\) 600
\(\times\) 1520 mm or 24 \(\times\) 24 \(\times\) 60 inches) indoor farming
solution specially designed to grow cannabis and other medicinal
plants. Leaf is more expensive (\$1500) than the AeroGarden
platform. However, Leaf provides the possibility to control and
adapt the environment around the plants. On the other hand, it is
still a closed platform (both hardware and software) and its
customizability is low.

\subsection{Grove}
\label{sec:orgheadline5}
\label{orgtarget6}

Grove offers a solution named ``The Garden''\footnote{\url{http://grovegrown.com}}. This
medium-size device (830 \(\times\) 400 \(\times\) 1900 mm or 33
\(\times\) 16 \(\times\) 75 inches) includes an aquarium to complement the
system (i.e., plants receive the nutrients from fish organic
material). The system comes with a smartphone application that can
track pH and bacteria levels, send reminders to the user, offer
growing tips, etc. However, this device cannot control or adapt the
environment around the plants since it does not rely on a sealed
chamber and its customizability is also low. The Garden's retail
price is higher than previous platforms (\$4500).

\subsection{Conviron A1000}
\label{sec:orgheadline6}
\label{orgtarget7}

The A1000\footnote{\url{http://www.conviron.com/products/a1000-reach-in-plant-growth-chamber}} platform is targeted towards small labs
and plant scientists interested in plant growth research. Even
though the primary customer base of the A1000 isn't the end-user as
previous platforms, Conviron chambers such as the A1000 have turned
into the standard for running plant experiments within the academic
community. Therefore, the comparison of its technology with other
devices provides important information. The A1000 is a large device
(1040 \(\times\) 825 \(\times\) 2020 mm or 41.75 \(\times\) 32.5 \(\times\)
79.5 inches) with the possibility of controlling environmental
variables like photoperiod, light intensity, air temperature,
humidity, \(CO_{2}\), etc. In spite of the controllable environment
capabilities, the A1000 is a closed platform with low
customizability. Its price is on request. However, it runs on the
order of tens of thousands of dollars, which makes it the most
expensive solution within our list.

\section{Robot design for personal environment-controlled applications}
\label{sec:orgheadline13}
\label{orgtarget8}

Most of the commercial devices mentioned above are exclusively
non-adaptive systems (i.e., do not allow the user to change the
environment around the plant) or they are based on closed systems
that provide very low customizability capabilities. Finally, the
relatively high prices of systems such as that proposed by Grove
Labs or Conviron hinder the adoption of this technology by end users
or research institutes. Therefore, a device that tries to overcome
these problems would need to satisfy the following criteria:

\subsection{Desktop size}
\label{sec:orgheadline8}

A controlled-environment device that can operate on a desk in an
indoor research environment (e.g., classroom) or inside a typical
apartment setting increases the possibilities of using this
technology and experimenting with different environmental
conditions. We consider that a suitable size might match the
specifications of traditional home appliances. Thus, the device
should not be bigger than 1000 \(mm^{3}\) or 39 3/8 \(inches^{3}\). 

\subsection{Wide range of possibilities}
\label{sec:orgheadline9}

In order to make this tool useful for diverse disciplines such as
plant phenotyping research or robotics academic curricula, the
proposed device should allow a broad range of possibilities in its
basic version. A modular design to which different sensors and
actuators can be added is a crucial aspect in order to customize
the system for a wide audience.

\subsection{User friendly}
\label{sec:orgheadline10}

The interfaces that allow users to communicate with and extract
information from the device must be intuitive, simple, and
efficient. These are important points in order to engage a wide
variety of users with the proposed system. 'Plug and play'
connections with the sensing, actuation, and computing parts of the
system need to be emphasized. Finally, the capability of creating,
storing, and sharing relevant data generated by the device is also
an important feature of the user interface.

\subsection{Low cost}
\label{sec:orgheadline11}

To assure the adoption of this technology by different types of
users (e.g., makers, food enthusiasts, researchers, etc.), the
proposed device needs to be affordable. We propose reducing the
cost of the system by using non-proprietary software and hardware
solutions.

\subsection{Open Information}
\label{sec:orgheadline12}

This platform needs to be shared among different types of users
with different requirements such as teachers, students, or
scientific staff. To provide a suitable platform to replicate
experiments and analyse the data obtained by the device, an
open-source hardware/software development model is an effective
approach.\\

None of the platforms outlined above and currently on the market
fulfills these criteria. Most controlled-environment devices are large
and thus need to operate in outdoor environments or special locations.
Also, very few are based on open standards. Due to the current gap in
the field, we propose the latest version (v2) of the PFC. The
following sections present the PFC v2 design.

\section{The OpenAg\texttrademark{} Personal Food Computer (PFC)}
\label{sec:orgheadline22}
\label{orgtarget2}

The PFC is an open-source open-hardware platform; its design
prioritizes the criteria mentioned above: desktop size, low cost,
customizability, user friendliness, and open information. These five
constraints imply the following actions: First, to reduce the size of
a system with a large number of components. Second, to reduce the
price of the device using cheap ``off-the-shelf'' components and mass
production manufacturing techniques. Third, to obtain a user-friendly
device by providing an intuitive and interactive user interface as
well as a modular hardware system to add or remove sensing, actuation,
or computing devices.

\begin{figure}[!tb]
\centering
\includegraphics[width=\columnwidth]{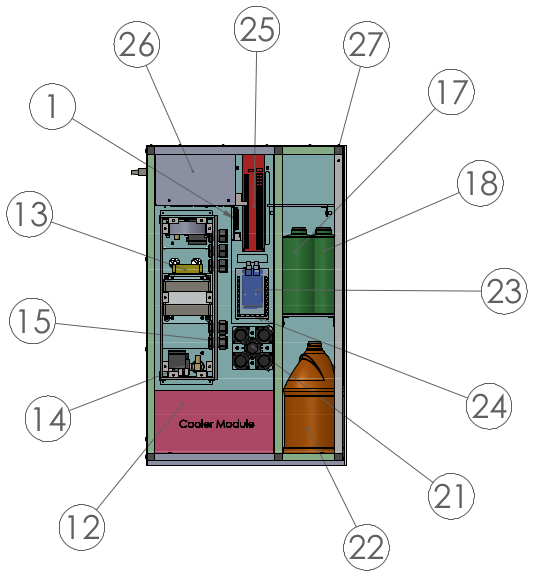}
\caption{\label{fig:orgparagraph3}
Side view of the PFC v2. The main electronic panel and its components are described.}
\end{figure}

\begin{figure}[!tb]
\centering
\includegraphics[width=\columnwidth]{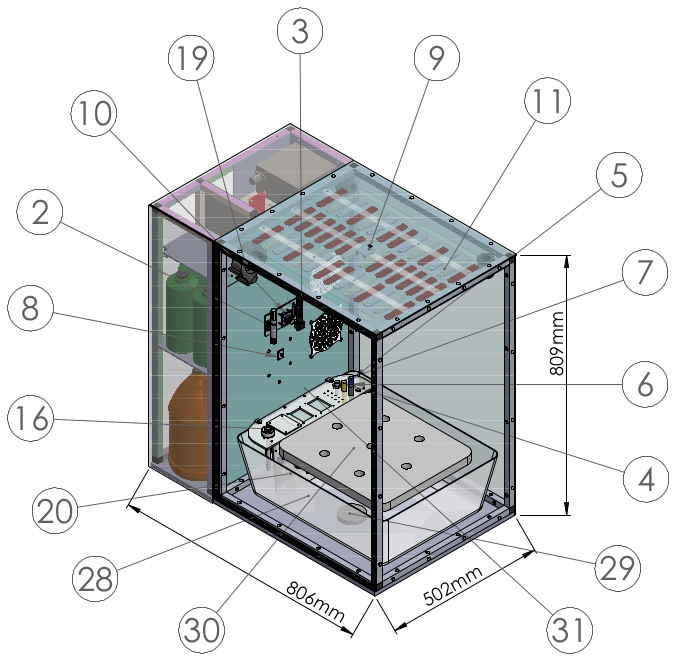}
\caption{\label{fig:orgparagraph4}
Orthogonal view of the PFC v2. Device measures and the main elements of the growing chamber are described.}
\end{figure}

In this design process, a central aspect is the choice of the PFC
capabilities. This particular choice is one of the innovations of the
PFC design. The sensors, actuators, and interfaces of the PFC
represent a wide range of devices one can find in several engineering
sub-domains. Fig. \ref{fig:orgparagraph3} and Fig. \ref{fig:orgparagraph4}
show the mechanical structure and the different hardware components
composing the PFC. Fig. \ref{fig:orgparagraph5} outlines the
connection diagram of these components. The following section explains
the hardware and software design choices in detail.

\subsection{The PFC Hardware}
\label{sec:orgheadline17}
\label{orgtarget9}

\subsubsection{Single Board Computer (SBC)}
\label{sec:orgheadline14}
\label{orgtarget10}

The electronic structure of the PFC is built around the
open-hardware platform Raspberry Pi 3\footnote{\url{http://www.raspberrypi.org/products/raspberry-pi-3-model-b/}} (depicted
in Fig. \ref{fig:orgparagraph3} \circled{1}). The Raspberry Pi has a
strong community of users. This SBC provides enough flexibility
because it embeds both multi-purpose ports such as Universal
Serial Bus (USB) as well as General Purpose Input/Output (GPIO)
ports, which allow the integration of more complex circuits. The
Raspberry Pi 3 CPU runs at A 1.2GHz 64-bit quad-core with an ARMv8
architecture. This architecture is supported by most of the Linux
distributions, allowing access to relevant tools such as different
programming languages and compilers, robotics-related software,
etc.

\subsubsection{Sensors and actuators}
\label{sec:orgheadline15}
\label{orgtarget11}

To allow a wide range of experimentation possibilities, the
PFC contains various sensing devices:

\begin{itemize}
\item The AM2315 sensor (\circled{2}) was used
as an air temperature and humidity sensor. This sensor has a
resolution of 0.1 units for both relative humidity (\%RH) and
temperature (\degree C). Its temperature sensing range is -40 to
125\degree C with a humidity repeatability of \(\pm\) 0.1 in \%RH.
These capabilities are sufficient to fulfill the needs of the
controlled-environment agricultural applications.

\item The MHZ16 sensor (\circled{3})  was used
as a \(CO_{2}\) sensor. Its \(CO_{2}\) detection range is 0-2000
ppm. Even though it requires up to 2-3 minutes to warm up before
reporting valid data, its Universal Asynchronous
Receiver/Transmitter (UART) connectivity provides an easy way to
set up the sensor.

\item The Atlas pH sensor (\circled{4}) was
chosen as a solution to measure the pH in the reservoir tub
\circled{28}. This sensor has a pH range of 0 - 14 (Na+ error at
> 12.3 pH). Its small size (12 mm \texttimes{} 150 mm) is ideal
to meet the constraints of our personal-size system.

\item Complementarily, the Atlas EC sensor (\circled{5}) was
used for measuring the electrical conductivity (EC). With
similar measures and connecting interfaces as the Atlas pH
sensor, it was easy to integrate it into our system. With a
conductivity range of 5 µS/cm to 200,000 µS/cm, it was more than
sufficient to meet the needs of an agricultural system.

\item The 1-Wire interface DS18B29 sensor (\circled{6})
was chosen as the water temperature sensing unit. With a \textpm{}
0.5\degree C accuracy from -10\degree C to +85\degree C and the
possibility to conduct underwater measurements, this device was
suitable for the PFC specifications.

\item A water level sensor (\circled{7}) provides the PFC
with the possibility to detect when the water in the bay needs
to be refilled. We chose the LLE102000 sensor,
which offers an accurate resolution of \textpm{} 1 mm.

\item The TSL2561 developed by Adafruit was chosen as the light
intensity sensor (\circled{8}). With an I2C
interface and light ranges from up to 0.1 - 40000+ Lux, the
TSL2561 provides enough resolution to meet our system's
specifications.

\item Two USB ELP 3.6mm Lens 5 Megapixel cameras are used
to obtain images from the plants and run computer vision
algorithms. The first one is located at the top of the PFC
(\circled{9}). The second one is located at the side of 
the box (\circled{10}).\\
\end{itemize}

Regarding the actuation side of the system, the following components
were selected:

\begin{itemize}
\item Grow Lights: GE Light modules (\circled{11}) with Red, Blue,
White (individually controllable / PWM) channels were chosen as
the main lighting actuator.

\item Air Cooler: A KippKitts cooling unit
(\circled{12}) with 200W was selected as the main chiller mechanism.

\item Air Heater: A 12V 150W electric ceramic thermostatic PTC
heating element (\circled{13}) was chosen as the main heating
element.

\item Fresh air valve: A 1/2 inch DC12V motorized ball (\circled{14})
valve was selected in order to manage the exchange of air
between the growth chamber and its exterior environment.

\item Cable gland: Four 3/4 inch NPT thread cable glands
(\circled{15}) were used as connecting points for the passing
wires and cables from and to the air manifold. In addition, they
are designed as modular air connection points for \(CO_{2}\)
dosing or other gas dosing (e.g., \(NO_{2}\), particulate manner,
aerosols, etc.).

\item Humidifier: Phtronics portable bottle cap air
humidifier with bottle (\circled{16}) was chosen as the main
source of humidity for our system.

\item pH and Nutrients solutions: Two liquid bottles (pH Up
\& pH Down) (\circled{17}) from
General Hydroponics (1-Quart) and two FloraDuo (A \&
B) (\circled{18}) fertilizers were chosen as
main solution units respectively.

\item Air circulation fan: A DC blower (\circled{19}) with
rated current of 1 Amp and voltage of 12V was used as a
circulation fan within the growing chamber.

\item Water circulation pump: A 12V submersible circulation
pump (\circled{20}) was chosen as the main device to activate
the water manifold.

\item Peristaltic Liquid Pumps: Five homecube 12V DC peristaltic
liquid pumps are located at the side of the PFC (\circled{21}).
Two of them are in charge of pumping the pH solutions
(\circled{17}) into the tub (\circled{28}). Another two are in
charge of pumping the nutrients into the tub. Finally, the
remaining one is in charge of pumping fresh water from the water
reservoir (\circled{22}).
\end{itemize}

In order to connect sensory input with actuation power, the following
electronic equipment was used:

\begin{figure}[!tbh]
\centering
\includegraphics[width=.9\linewidth]{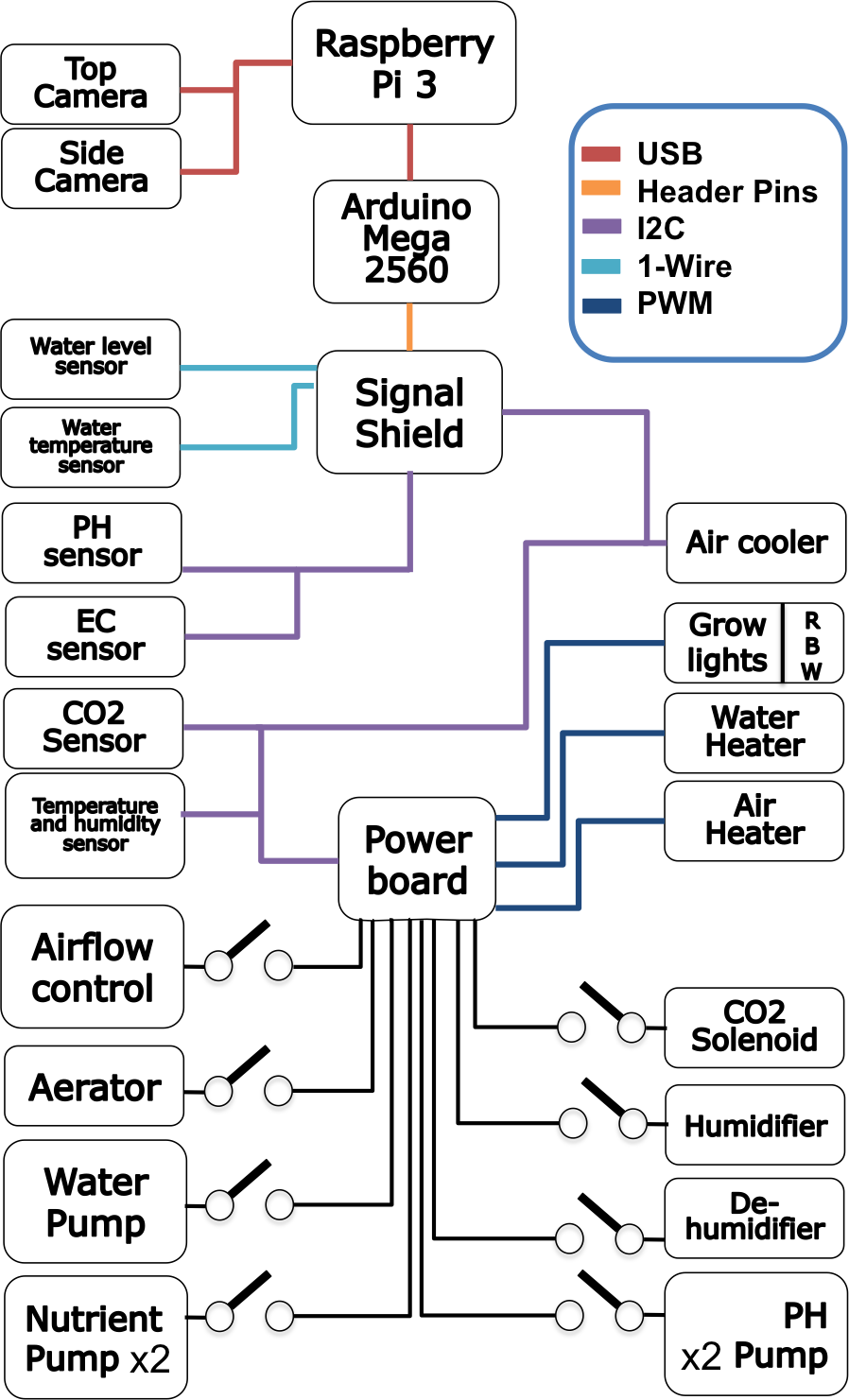}
\caption{\label{fig:orgparagraph5}
Electronics and connection diagram for the PFC.}
\end{figure}

\begin{itemize}
\item Arduino Mega 2560: This board (\circled{23}) was used in order
to easily add and remove the low-level sensors as well as to
make the most of the code and community already built around
this family of boards.

\item Signal shield: The SainSmart Sensor Shield v2 (\circled{24}) was
chosen as the main signal board. This shield is directly mounted on
top of the Arduino Mega 2560 and its main functionality is to
read from the multiple sensors of the system and send commands
to the power board.

\item Power board: A customized PCB (\circled{25})
board composed of MOSFETS \& Relays electronic components was
used for high power switching. This board has a standard
interface that can be controlled directly by the Arduino Mega
2560 board.

\item Power Supply: A 500W power supply unit (\circled{26}) is in
charge of powering the whole system.
\end{itemize}

\subsubsection{Mechanics}
\label{sec:orgheadline16}
\label{orgtarget12}

\begin{itemize}
\item Structural Frame: The outer frame (\circled{27}) is a 806 \texttimes{}
502 \texttimes{} 809 mm (31.7 \texttimes{} 19.7 \texttimes{} 31.8 inches) structure
composed of 3/4 inch (19 mm) anodized aluminum extrusion tubing
connected by push-in brackets that assemble different edges of
the frame.

\item Reservoir Tub: The purpose of the reservoir tub (\circled{28})
is to hold nutrient-rich water and to keep 
the root zone dark to reduce algae growth. The water in the tank
will do the same job that soil does when plants are grown
outdoors.

\item Aerator, tubing, and stone: (\circled{29}) Plant roots
need oxygen in the water for respiration. This system increases
the dissolved oxygen content of the water in the reservoir
tub as well as agitates the water to prevent root mold and
disease.

\item Styrofoam float tray: This board is the structural anchor where the
seedlings are planted (\circled{30}). The foam tray is like a
raft that floats on top of the nutrient rich water, creating a
barrier between the leaf zone and the root zone. This piece also
helps to prevent algae growth by blocking light to the root
zone.

\item Outer Shell: The shell (\circled{31}) is what separates
the internal climate of the growth chamber from the outside
environment. This part of the PFC provides a physical barrier
and styrofoam insulation to help maintaining internal
conditions. The shell has a transparent window so users can
observe their plants without disturbing the inner climate. \\
\end{itemize}

Detailed information about every sensor, actuator, and mechanical part
used in the system, can be found in a BOM (Bill Of Materials)
file\footnote{\url{http://goo.gl/7zfGsB}} provided within the Wiki page of the project\footnote{\url{http://wiki.openag.media.mit.edu/}}. In
this extensive list, information such as part supplier, description,
cost, and datasheet links are provided. Finally, assembly instructions
and a complete list of building resources can be found within the
official PFC (v2) repository\footnote{\url{http://github.com/OpenAgInitiative/openag_pfc2}}.

\subsection{The PFC software}
\label{sec:orgheadline21}
\label{orgtarget13}

\subsubsection{User interface}
\label{sec:orgheadline18}

\begin{figure}[!tbh]
\centering
\includegraphics[width=\columnwidth]{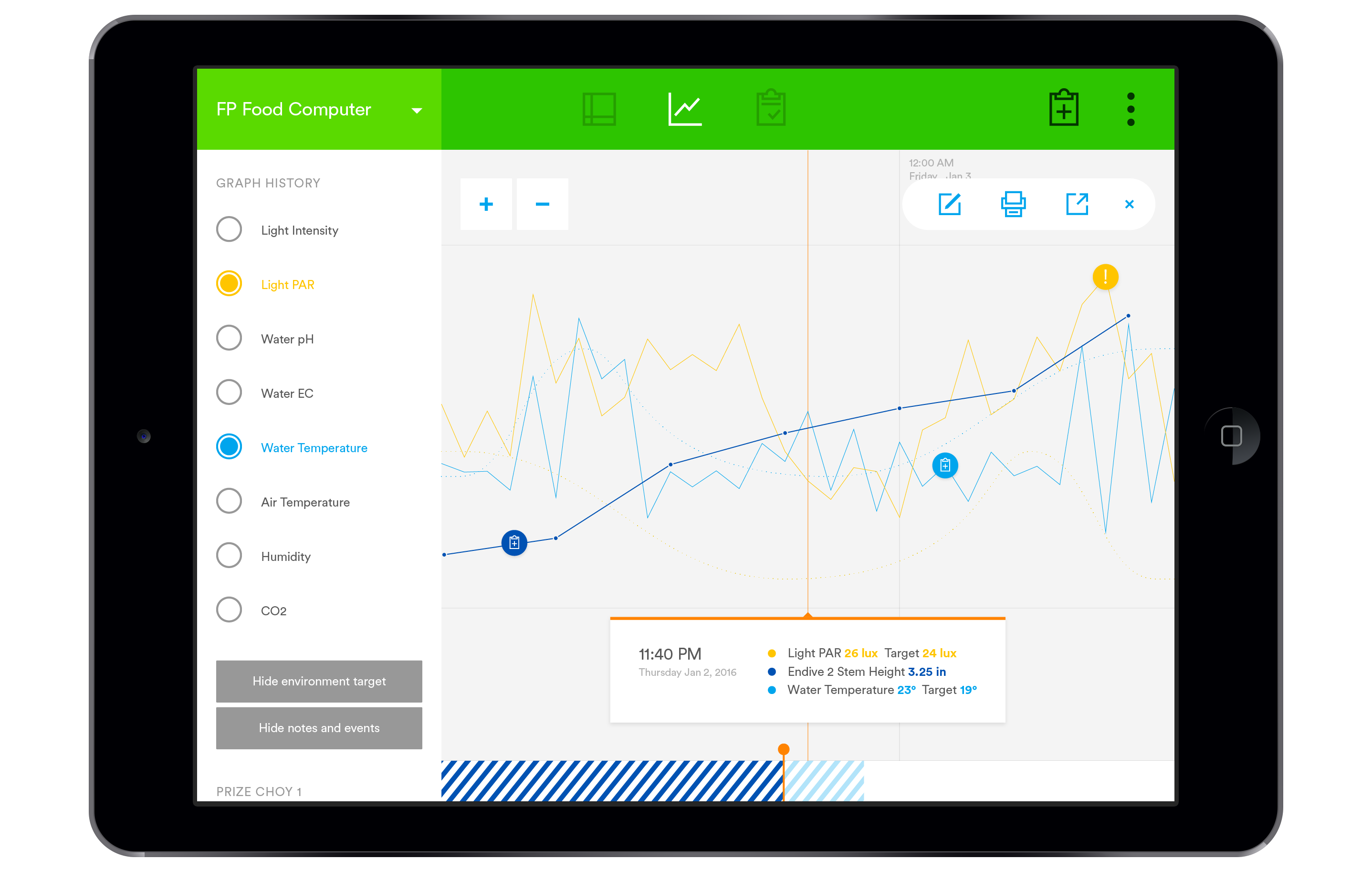}
\caption{\label{fig:orgparagraph6}
Screenshot of the User Interface (UI) where a sequence of environmental sensor data points (a.k.a. ``climate recipe'') is visualized. This feature of the UI provides the possibility to change the time scale of the recipe as well as bookmark certain parts of it for future reference.}
\end{figure}

\begin{figure}[!tbh]
\centering
\includegraphics[width=\columnwidth]{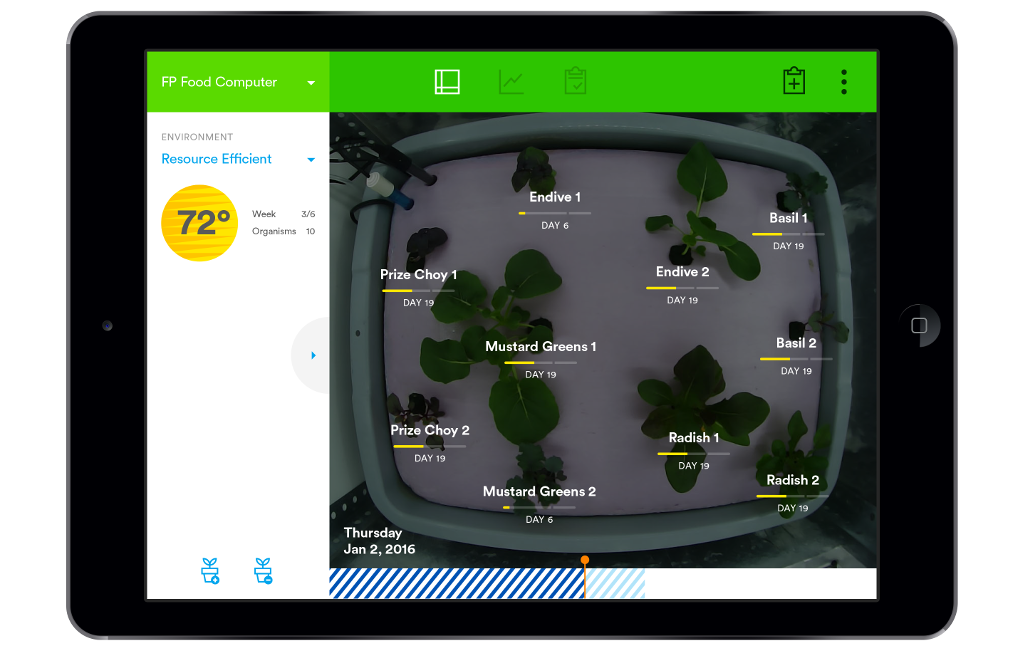}
\caption{\label{fig:orgparagraph7}
Screenshot of the UI where an image of the growing chamber is displayed. This feature of the UI allows the user to create time-lapse videos and monitor plant growth with progression bars.}
\end{figure}

In order to provide a user-friendly environment to operate the
PFC, we developed a web-based interface using the Javascript
language. Several features of this UI include:

\begin{itemize}
\item Visualization of environmental data points and progress of plant
growth (Fig. \ref{fig:orgparagraph6}).
\item Loading, modifying, and exporting climate recipes and output
data in CSV format.
\item Creation of time-lapse videos and visualization of camera feeds
(Fig. \ref{fig:orgparagraph7}).
\item Manually actuating different devices in the PFC such as pumps,
heater, cooler, etc.
\end{itemize}

The UI is designed to run on any device with a web browser (e.g.,
tablets, mobile phones, desktop computers, etc.). This feature
allows the user to operate the PFC from a proximity scenario using
the local WiFi network or a remote scenario using the internet.
This capability increases the flexibility of the PFC users to
monitor their experiments. The source code of the UI can be found
here\footnote{\url{http://github.com/OpenAgInitiative/openag_ui}}.

\subsubsection{Climate Recipes}
\label{sec:orgheadline19}

As commented before, climate recipes embedded a sequence of
environmental sensor data points that represent climates where an
organism is grown. Currently, the PFC only supports one ``simple''
recipe format. However, the system is designed to allow new
formats to be developed over time. This ``simple'' recipe format
conceptualizes recipes as a sequential list of set points for
environmental variables. In particular, a “simple” recipe is a
list of 3-element lists with the following structure:

\phantomsection
\label{orgexampleblock1}
\begin{verbatim}
[<offset>,    <variable_type>,    <value>]
\end{verbatim}

Where \texttt{<offset>} is the number of seconds since the start of the
recipe at which this set point should take effect,
\texttt{<variable\_type>} is the variable type to which the set point
refers (e.g. ``\texttt{air\_temperature}''), and \texttt{<value>} is the value of
the set point. The set point stays in effect until a new set point
for that variable type is reached. The list of set points is
ordered by offset. The recipe will end as soon as the last set
point is emitted. Recipes are encoded into JSON files that are
inputed into the PFC control system. A sample of this simple
recipe format can be depicted in the following example:

\phantomsection
\label{orgexampleblock2}
\begin{verbatim}
{"_id": "7ca3134e91aec96acd17a74764000bb8",
"format": "simple",
"operations": [
    [0, "air_temperature", 25],
    [0, "air_humidity", 25],
    [0, "light_illuminance", 60],
    [43200, "air_temperature", 23],
    [108000, "light_illuminance", 0],
    [172800, "air_humidity", 20],
    .
    .
    .
}
\end{verbatim}

\subsubsection{On-board software structure}
\label{sec:orgheadline20}

\begin{figure*}[!tbh]
\centering
\includegraphics[width=0.95\textwidth]{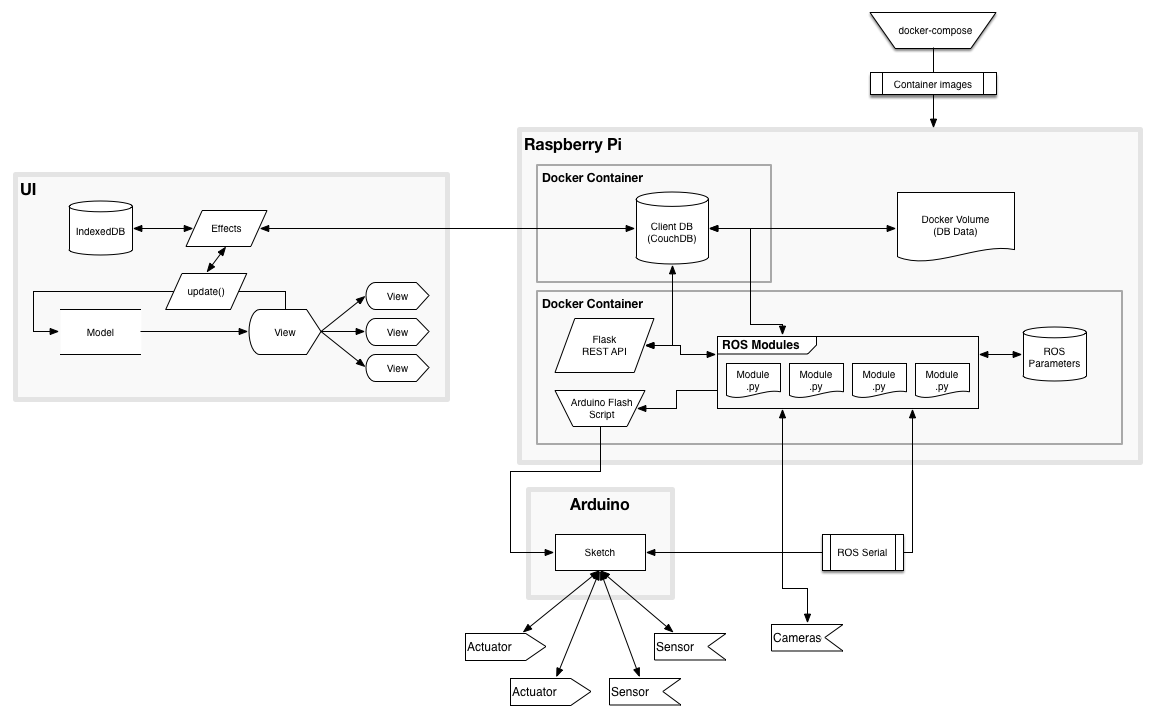}
\caption{\label{fig:orgparagraph8}
Layout of the on-board software structure of the PFC. Two Docker containers are responsible for the whole system setup and configuration. The first container starts and loads a local CouchDB instance. This NoSQL database stores all sensor data and previous climate recipes. The second container is responsible for setting up the ROS infrastructure and applying the control algorithms for climate recipe handling.}
\end{figure*}

Fig. \ref{fig:orgparagraph8} outlines the PFC's on-board
software structure and its interaction with external components
such as the UI. A NoSQL database (CouchDB) was chosen to store the
data coming from the PFC sensors as well as previous completed
recipes. The UI uses this local database to visualize the current
sensor values and monitor the progression of the current climate
recipe. 

To control the PFC hardware we decided to use ROS \cite{Quigley2009}
(Robot Operating System). We made that choice to reuse already
developed control schemes (e.g. PID controllers), sensing packages,
debug tools, etc. and to be able to interface with other robots in
the future. Within our ROS infrastructure different ROS modules are
in charge of controlling the PFC. For instance, one module loads
and stores information from and to the local CouchDB database
(Client DB), another flashes the Arduino board to provide different
sensor capabilities or communicates with the Flask REST API to
provide HTTP control points to the whole system. The behavior of
these modules can be changed online by the use of different ROS
parameters. A detailed list of ROS modules and its parameters can
be found within the \texttt{openag\_brain} package\footnote{\url{http://github.com/OpenAgInitiative/openag_brain}}. Two
Docker containers are used to accomplish this whole setup. Docker
containers allow us to package the software to be reproducible,
easy to deploy, and ready to operate the PFC. Docker containers are
stateless, which means that to create a certain configuration data
needs to be saved outside the containers (Docker Volume).

\section{PFC Design Challenges}
\label{sec:orgheadline26}
\label{orgtarget14}

During the design phase of the PFC, we faced several challenges in
order to meet the criteria introduced in section
\ref{orgtarget8}. We describe three of these challenges in the
following lines:

\subsection{Hardware agnostic control}
\label{sec:orgheadline23}

Due to the high-customizability capabilities of a system such as
the PFC, we realised that a wide range of hardware components might
be added, removed, substituted, or sourced locally by end users. In
order to cope with this hardware variability, we decided to program
the software stack of the PFC with hardware agnostic orders (e.g.,
``dose 20 ml of solution'') rather than component specific commands
(e.g., ``turn the pump on for 2 secs''). Using this approach we can
achieve feedback control models like the one described in Fig.
\ref{fig:orgparagraph1} without specific or specialized hardware
components. This hardware agnostic control represents an
improvement from previous controlled-environment platforms (e.g.,
Conviron A1000\footnotemark[7]{}, Leaf\footnotemark[5]{}, etc.), which
heavily depend on explicit hardware.

\subsection{Frame modularity and scalability}
\label{sec:orgheadline24}

The PFC is designed as a desktop size system. However, major crops
such as corn, wheat, or soy (of great interest to the scientific
community) require bigger chambers to grow. The PFC's structural
frame is made out of standardized aluminum extrusion tubing
connected by push-in brackets. This structural approach was chosen
to allow the user to expand the growth chamber horizontally and
vertically (depending the needs of the plant or crop) maintaining
the same sensing, actuation and control components. The assembly of
the different edges as well as the expansion of the frame to bigger
size chambers requires minimal changes and no specialised tooling.
The possibility of scaling the PFC to accommodate different types
of plants and crops represents an improvement from previous
controlled-environment platforms, which are unable to adapt their
growing space.

\subsection{Open Database for plant growth data}
\label{sec:orgheadline25}

To provide a suitable platform to replicate experiments and analyse
the data obtained, we envision an open database for plant growth
data\footnote{\url{http://www.media.mit.edu/research/groups/open-phenome-project}}, where controlled-environment devices
such as the PFC act as end-effectors. For this purpose, we decided
to implement a client/server database architecture, where the PFC
local database (Client DB) can operate offline while a cloud
database instance (Server DB) can be synchronized once the PFC
comes online. A filtered replication method between both database
instances has been developed to save bandwidth and space on the
client side. Client databases only download the changes that take
place in the cloud database. However, local PFC databases upload
all their content to the cloud.

\section{Previous deployments}
\label{sec:orgheadline29}
\label{orgtarget15}

\subsection{Evaluation of the PFC by teachers}
\label{sec:orgheadline27}

In September 2015, six high schools in Massachusetts were selected
to conduct a pilot program to use PFCs in classrooms. A total of
200 students (30 students per 6 classrooms) were engaged in
building and programming a PFC to create academic curricula.

\begin{figure}[!tbh]
\centering
\includegraphics[width=\columnwidth]{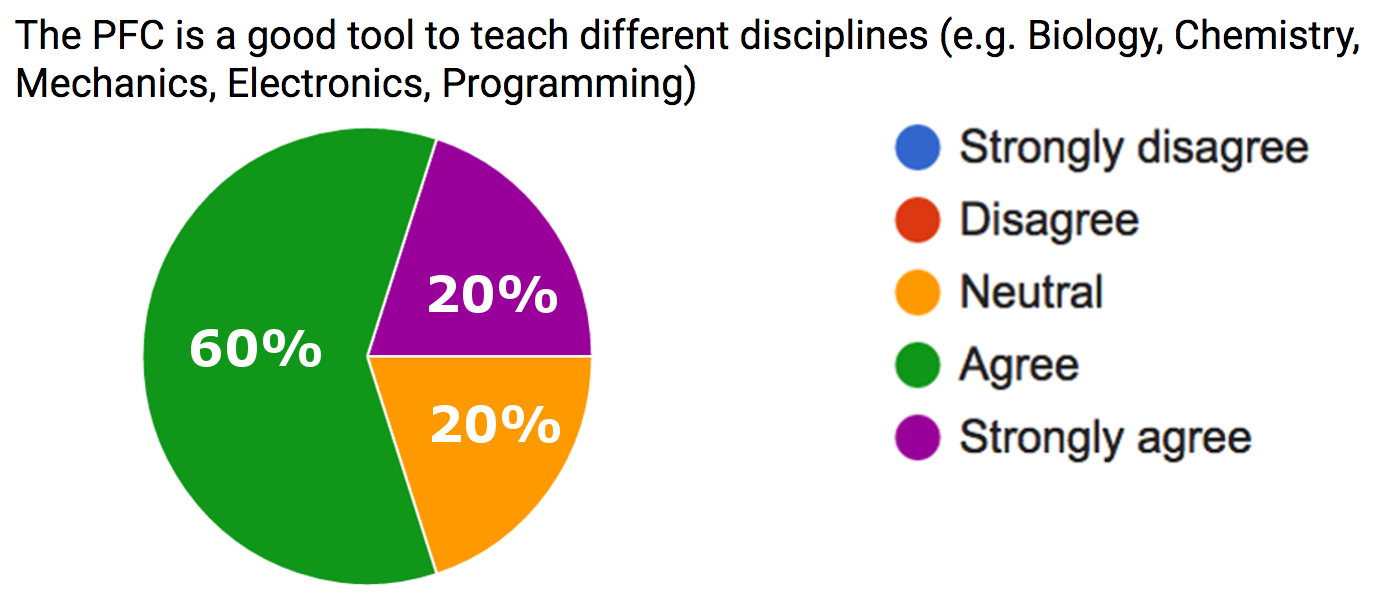}
\caption{\label{fig:orgparagraph9}
Results depicted from the survey distributed to 5 teachers during the pilot program.}
\end{figure}

After this pilot program, teachers were asked to give feedback
about the use of the PFC and its capabilities to create curricula
in different disciplines such as biology, chemistry, programming,
electronics, etc. Fig. \ref{fig:orgparagraph9} depicts the results of
this survey and shows that more than 80\% of the teachers agree or
strongly agree that the PFC is a good tool to teach different
disciplines and generate academic curricula.

As a result of this pilot program, an educator’s user guide and
associated curricula was created \cite{Rogoff2016}. The research that
lead to this user's guide suggests that the PFC introduces an
opportunity for students to grow their food with an exciting and
fun tool that they can take ownership of. The PFC also introduces
opportunities for students to engage in exciting, cutting-edge
technologies. This pilot program led to a 2016 Edison Award (Bronze
Category)\footnote{\url{http://www.edisonawards.com/winners2016.php}}.

\subsection{User Community}
\label{sec:orgheadline28}
\label{orgtarget16}

\begin{figure}[!tbh]
\centering
\includegraphics[width=\columnwidth]{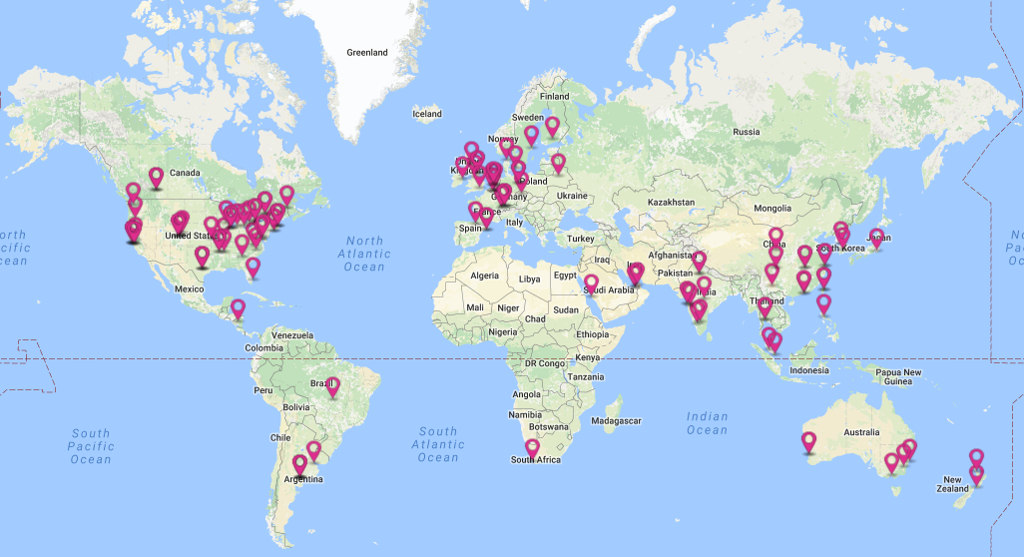}
\caption{\label{fig:orgparagraph10}
Locations where different PFCs (v1 or v2) are being built.}
\end{figure}

We created an open online forum\footnote{\url{http://forum.openag.media.mit.edu}} and a Wiki page to
disseminate the research progress on the PFC project. Moreover, we
are in the preliminary stages of providing non-English translations
of our documentation. Since launching the OpenAg community in May
2016, we gathered a total of 855 users from every continent, 297
topic threads, and about 3500 posts. Currently, different PFCs are
being built in 33 countries around the world (Fig. \ref{fig:orgparagraph10}).

\begin{figure}[!tbh]
\centering
\includegraphics[width=\columnwidth]{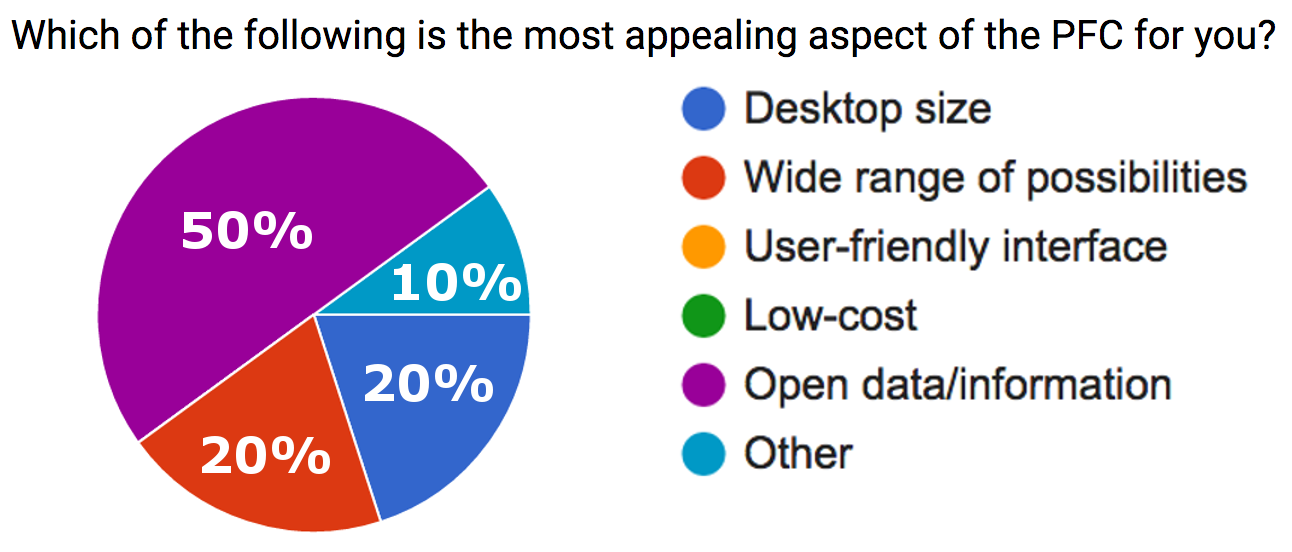}
\caption{\label{fig:orgparagraph11}
Results depicted from the survey distributed to 20 users which recently completed the PFC building process.}
\end{figure}

Another survey was distributed to a limited group of users which
recently completed the PFC building process. These users were asked
about what were the most appealing aspects of the PFC for them.
Fig. \ref{fig:orgparagraph11} shows that 90\% of the users
find PFC aspects such as the open information, wide range of
possibilities (i.e., customizability), or its desktop size
appealing in order to build and use this kind of solution. These
results correlate to our design premises described at the beginning
of this paper.

\section{Availability}
\label{sec:orgheadline30}
\label{orgtarget17}

The first kits (hardware and building instructions) for the PFC (v2)
were ready for shipment by early 2017. However, all software (e.g.,
source code and development tools) and hardware components (e.g.,
CAD drawings) are accessible online.

\section{Conclusions}
\label{sec:orgheadline31}
\label{orgtarget18}

The PFC is an innovative personal controlled-environment device for
growing plants. However, it is also useful to teach a wide range of
topics. For its size, price, and capabilities the PFC is a complex
system that can be used not only as a research platform but also as
an educational tool. The open-source nature of the PFC improves the
quality of the support for the end users by providing full access to
knowledge at every level. This paper presents the main hardware,
software and design components behind the PFC as well as provides
information about previous deployments and evaluations of the
proposed platform. For future work, we envision an open-source
digital library for plant growth data, where controlled-environment
devices such as the PFC act as distributed data collecting stations.

\section{Acknowledgements}
\label{sec:orgheadline32}

We thank MIT Media Lab, Target Corporation, Welspun, LKK, Google,
Pentair and Partners in Food Solutions for funding the development
of the PFC project.

\bibliographystyle{IEEEtran}
\bibliography{References}
\end{document}